\pdfoutput=1
\documentclass[conference]{IEEEtran}
\IEEEoverridecommandlockouts
\usepackage{cite}
\usepackage{amsmath,amssymb,amsfonts}
\usepackage{algorithmic}
\usepackage{graphicx}
\usepackage{textcomp}
\usepackage{xcolor}
\usepackage{subcaption}
\usepackage{multirow}
\usepackage{booktabs}
\def\BibTeX{{\rm B\kern-.05em{\sc i\kern-.025em b}\kern-.08em
    T\kern-.1667em\lower.7ex\hbox{E}\kern-.125emX}}
\begin{document}

\title{Compressing Facial Makeup Transfer Networks by Collaborative Distillation and Kernel Decomposition \\
\thanks{*Haoji Hu is the corresponding author of this paper.}
}
\author{\IEEEauthorblockN{Bianjiang Yang$^1$, Zi Hui$^2$, Haoji Hu$^2$*, Xinyi Hu$^2$, Lu Yu$^2$}
\IEEEauthorblockA{$^1$\textit{Chu Kochen Honors College},
\textit{Zhejiang University, Hangzhou, China}\\
$^2$\textit{College of Information Science and Electronic Engineering},
\textit{Zhejiang University, Hangzhou, China}\\
}
\textit{Emails: \{yangbj, zihui, haoji\_hu, xinyih, yul\}@zju.edu.cn}}

\maketitle

\begin{abstract}
Although the facial makeup transfer network has achieved high-quality performance in generating perceptually pleasing makeup images, its capability is still restricted by the massive computation and storage of the network architecture. 
We address this issue by compressing facial makeup transfer networks with collaborative distillation and  kernel decomposition.
The main idea of collaborative distillation is underpinned by a finding that the encoder-decoder pairs construct an exclusive collaborative relationship, which is regarded as a new kind of knowledge for low-level vision tasks. 
For kernel decomposition, we apply the depth-wise separation of convolutional kernels to build a light-weighted Convolutional Neural Network (CNN) from the original network. Extensive experiments show the effectiveness of the compression method when applied to the state-of-the-art facial makeup transfer network -- BeautyGAN~\cite{LiQiaDonLiuYanZhuLin18}. 
\end{abstract}


\section{Introduction}
Facial makeup transfer aims to translate the makeup style from a given reference face image to another non-makeup face image without the change of face identity. It is an interesting but challenging area because it needs to handle different local styles/cosmetics including eye shadow, lipstick and foundation in a coherent and natural way. Existing research works on automatic makeup transfer can be classified into two categories: traditional image processing approaches such as physics-based manipulation~\cite{LiZhoLin15,TonTanBroXu07}, and deep learning based methods which typically build upon deep neural networks~\cite{LiuEtal16,LiQiaDonLiuYanZhuLin18}.
A recent deep learning based network, BeautyGAN~\cite{LiQiaDonLiuYanZhuLin18}, further borrows the idea of style transfer networks~\cite{GatEckHerShe17,JohAlaFei16} 
and implements the encoder-decoder architecture for makeup transfer. 

Although facial makeup transfer network has achieved good performance, its capability is still restricted by the massive computation and high storage of the network architecture, which constrains its application on mobile devices. Model compression and acceleration methods provide a possible solution to save computation and storage of deep neural networks, which include parameter pruning~\cite{PenWuCheJun19}, quantization~\cite{CouHubEtAl16}, and knowledge distillation~\cite{HinVinDea15}, etc.
However, most compression methods only focus on high-level tasks, e.g., classification and detection. Compressing models for low-level vision tasks, such as facial makeup transfer, is still less explored.  

In this paper, taking BeautyGAN~\cite{LiQiaDonLiuYanZhuLin18} as an example, we address the problem of compressing facial makeup transfer networks to save computation and storage. Firstly, we apply Collaborative Knowledge Distillation (CKD)~\cite{WanLiWanHuYan20} to compress the encoder of BeautyGAN. 
The main idea of CKD is underpinned by a finding that the encoder-decoder pairs construct an exclusive collaborative relationship, which is regarded as a new kind of knowledge for low-level vision tasks. Thus, to overcome the feature size mismatch when applying collaborative distillation, a feature loss is introduced to drive the student network to learn a linear embedding of the teacher’s features.
Secondly, because BeautyGAN is not a standard encoder-decoder based network, it contains several residual blocks which are difficult to distill by CKD. 
To further reduce computation, we use the idea of kernel decomposition, which is proposed by MobileNets~\cite{HowZhuCheKaWanWeiAda17}.
Kernel decomposition are based on a streamlined architecture that uses depth-wise separable convolutions to build light-weighted deep neural networks.
Although extensive researches have shown the effectiveness of kernel decomposition in high-level vision tasks~\cite{HowZhuCheKaWanWeiAda17,Oso18}, it is rarely implemented in low-level tasks such as facial makeup transfer. We decompose the residual blocks of BeautyGAN into a depth-wise convolution followed by a~$1\times1$ point-wise convolution, thus reducing the computation by a large margin.

The major contributions of this paper lie in three aspects:
\begin{itemize}
\item We introduce an efficient two-step compression method for BeautyGAN by compressing the encoder with CKD and the residual blocks with kernel decomposition. 
\item We successfully demonstrate the effectiveness of kernel decomposition in facial makeup transfer network, indicating its usability in low-level vision tasks. 
\item Experiments on the MT dataset~\cite{LiQiaDonLiuYanZhuLin18} demonstrate that the compressed network performs favorably against the original BeautyGAN.
\end{itemize}

\section{The Proposed Method}
\begin{figure*}[htbp]
\centerline{\includegraphics[scale=0.7]{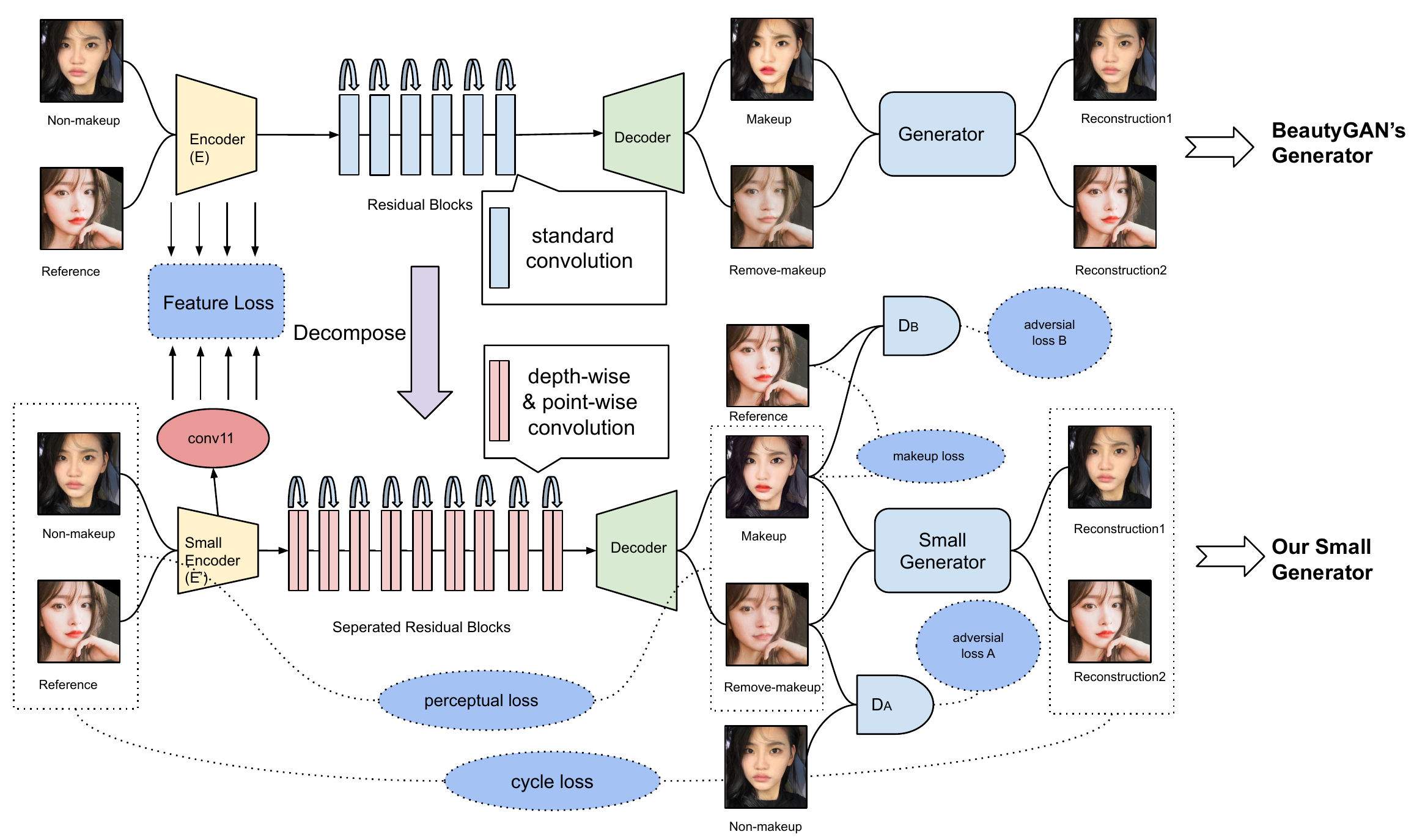}}
\caption{Overall framework of the proposed method.} 
\label{fig:framework}
\end{figure*}

The overall framework of the proposed method is illustrated in Figure~\ref{fig:framework}.
The original BeautyGAN at the top is the teacher network which sequentially consists of an encoder, $6$ residual blocks and a decoder. The student network is illustrated at the bottom which consists of a smaller encoder, $9$ decomposed residual blocks and a decoder. The four loss functions in BeautyGAN, i.e., adversarial loss, cycle loss, perceptual loss and makeup loss,  are also adopted by the student network. Collaborative distillation is implemented to distill the knowledge of the teacher's encoder to the encoder of the student network. The distillation process is guided by the feature loss between the two encoders.

Detailed architectures of the original BeautyGAN and our proposed student network are summarized in Table~\ref{tab: detail}.
The protocol of layer representation is `layer type-kernel size-number of input channels-number of output channels (stride)'. In addition, conv, res, deconv and separated\_res represent the convolutional layer, residual block layer, deconvolutional layer and separated residual blocks, respectively.

\begin{table}[h]
\setlength{\tabcolsep}{1.5mm}{
\caption{The architectures of the original BeautyGAN and our proposed student network.}
\centering
\begin{tabular}{|cc|cc|}\hline
\multicolumn{2}{|c|}{\textbf{BeautyGAN}}&\multicolumn{2}{c|}{\textbf{The Student Network}}\\\hline
\textbf{Non-makeup}&\textbf{Reference}&\textbf{Non-makeup}&\textbf{Reference}\\\hline
conv7-3-64(1)&conv7-3-64(1)&conv7-3-16(1)&conv7-3-16(1)\\  \hline
\multirow{2}*{conv4-64-128(2)}&\multirow{2}*{conv4-64-128(2)}&conv4-16-32(2)&conv4-16-32(2)\\
~&~&conv1-32-128(1)&conv1-32-128(1)\\\hline
\multicolumn{2}{|c|}{\textbf{Merge}}& \multicolumn{2}{c|}{\textbf{Merge}}\\ \hline
\multicolumn{2}{|c|}{conv4-256-256(2)}&\multicolumn{2}{c|}{conv4-256-256(2)}\\
\multicolumn{2}{|c|}{6 $\times$ res3-256-256(1)}&\multicolumn{2}{c|}{9 $\times$ seperated\_res3-256-256(1)}\\
\multicolumn{2}{|c|}{deconv4-256-128(2)}&\multicolumn{2}{c|}{deconv4-256-128(2)}\\
\multicolumn{2}{|c|}{deconv4-128-64(2)}&\multicolumn{2}{c|}{deconv4-128-64(2)}\\ \hline
\multicolumn{2}{|c|}{\textbf{Separate}}&\multicolumn{2}{c|}{\textbf{Separate}}\\\hline
\textbf{Makeup}&\textbf{Remove-makeup}&\textbf{Makeup}&\textbf{Remove-makeup}\\\hline
conv3-64-64(1)&conv3-64-64(1)&conv3-64-64(1)&conv3-64-64(1)\\
conv3-64-64(1)&conv3-64-64(1)&conv3-64-64(1)&conv3-64-64(1)\\
conv3-64-3(1)&conv3-64-3(1)&conv3-64-3(1)&conv3-64-3(1)\\\hline
\end{tabular}\label{tab: detail}}
\end{table}

\subsection{The Loss Functions}
As shown in Figure~\ref{fig:framework}, the same with BeautyGAN, we implement the four losses, i.e., the adversarial loss~$\mathcal{L}_{adv}$, cycle loss~$\mathcal{L}_{cyc}$, perceptual loss~$\mathcal{L}_{per}$ and makeup loss~$\mathcal{L}_{makeup}$,  to train the proposed student network. Please refer to~\cite{LiQiaDonLiuYanZhuLin18} for details of these losses.

\subsection{Collaborative Knowledge Distillation}
In this paper, we distill the encoder of BeautyGAN by collaborative knowledge distillation (CKD) which was originally proposed in style transfer network compression~\cite{WanLiWanHuYan20}.
BeautyGAN is an encoder-resnet-decoder based network, since the knowledge of the encoder is leaked into the decoder, we can compress the original encoder~$E$ to the small encoder~$E’$ by reducing the number of filters for each layer but maintaining the same architecture. Since the output dimensions of~$E'$ and~$E$ are different, CKD solves this mismatch problem by adding a $1\times1$ convolutional layer with linear activation function to each feature map of~$E'$. Suppose that the~$i$th feature of ~$E$ is represented as~$F_i$ and its counterpart of~$E'$ is represented as~$F_i'$. Basically the dimension of~$F_i'$ is smaller than that of~$F_i$. Then CKD proposes a \emph{feature loss} as follows:
\begin{equation}
L_{feat} = \sum_{i=1}^{n} || F_i’ \times Q_i – F_i ||_2 \label{eq_2}.
\end{equation}
where each~$Q_i$ is a learnable matrix to make the dimension of~$F_i'$ match the dimension of~$F_i$. 
 For our implementation, $n=4$ because we distill $2$ convolutional layers for both the non-makeup and reference branches. As shown in Table~\ref{tab: detail}, the first convolutional layers of the teacher and student's encoders are `conv7-3-64(1)' (BeautyGAN) and `conv7-3-16(1)' (student network), so the dimension of~$Q_i$ ($i=1,2$) is $16\times64$ to match the outputs of the convolutional layers of BeautyGAN and student network. Similarly, the second convolutional layers of the teacher and student's encoders are `conv4-64-128(2)' (BeautyGAN) and `conv4-16-32(2)' (student network), so the dimension of~$Q_i$ ($i=3,4$) is $32\times128$. 

The final loss function is the weighted combination of the feature loss and the four losses of BeautyGAN:
\begin{equation}
L = L_{feat}+\alpha L_{adv} + \beta  L_{cyc} +\gamma L_{per} + \sigma L_{makeup}.
\label{eq_4}
\end{equation}

\subsection{Residual Blocks Decomposition}
\begin{figure}[htbp]
\centerline{\includegraphics[scale=0.55]{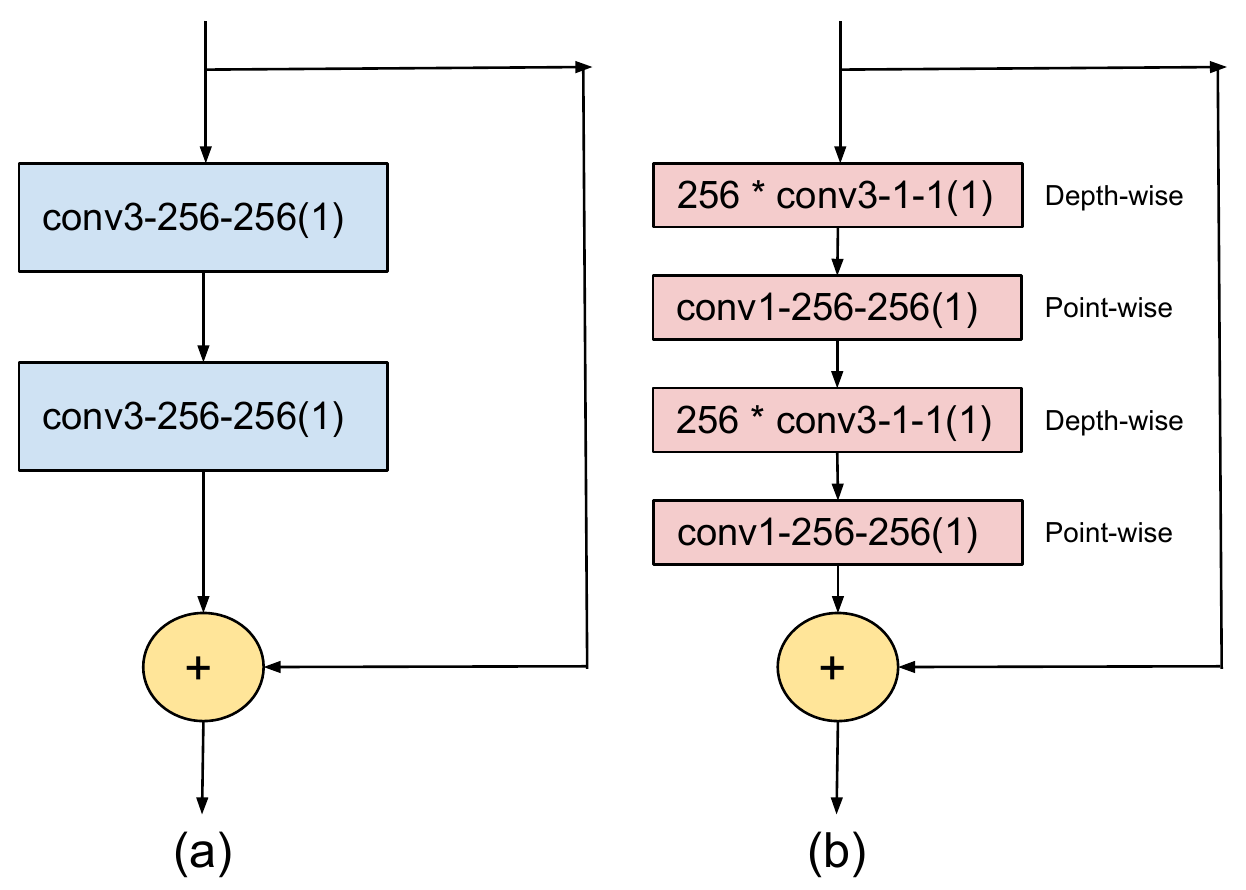}}
\caption{The residual block architectures. (a) The standard residual block of BeautyGAN; (b) The decomposed residual block of the student network.}
\label{fig:resnet decompose}
\end{figure}

Following the idea of MobileNets~\cite{HowZhuCheKaWanWeiAda17}, we decompose the standard convolution of residual blocks in BeautyGAN into the combination of depth-wise and point-wise convolutions.
Figure~\ref{fig:resnet decompose}(a) and (b) show the architectures of the standard residual block of BeautyGAN and the decomposed block of the proposed student network, respectively.
We decompose the standard convolution of residual blocks into a depth-wise convolution which shares the same convolutional kernel for each of the input feature maps,  and followed by a $1\times1$ point-wise convolution to make the number of the output channels consistent with the original. Suppose that a standard convolution takes an $H_i\times W_i\times C_i$ feature map as input and outputs an $H_o\times W_o\times C_o$ feature map, where~$H$, $W$ and~$C$ refer to the height, width and channel of the feature maps, respectively. If the kernel size is~$K \times K$, the computation cost of a standard convolutional layer is
$C_{stand} = K^2C_i C_o H_i W_i$. After decomposition, the computation cost is changed to~$C_{decom} = K^2 C_i H_i W_i + C_i  C_o   H_i  W_i$, which is the sum of the depth-wise and point-wise convolutions.
Thus, the computation reduction~$R$ is calculated as:
\begin{equation}
\begin{split}
R = \frac{N_{decom}C_{decom}}{N_{stand}C_{stand}}
= \frac{N_{decom}}{N_{stand}}\left(\frac{1}{K^2}  + \frac{1}{C_o} \right)
\label{eq_7}
\end{split}
\end{equation}
Here, $N_{stand}$ and~$N_{decom}$ refer to the number of standard and decomposed residual blocks in the teacher and student networks respectively. We set~$N_{stand}=6$ for BeautyGAN and~$N_{decom}=9$ for the student network (Refer to Section~\ref{sec:experiments} for ablation study of~$N_{decom}$). If we put  $K = 3$ and~$C_o = 256$ into Equation~(\ref{eq_7}), we can obtain~$R = 0.173$, which means that the computation of the decomposed residual blocks has been reduced to~$17.3\%$ of the uncompressed ones.

\section{Experiments}
\label{sec:experiments}

Data for experiments is from the MT dataset provided in BeautyGAN~\cite{LiQiaDonLiuYanZhuLin18}, which consists of~$3,834$ high-resolution face images with and without makeups. We use~$3,600$ images for training and~$234$ images for testing.  The parameters in Equation~(\ref{eq_4}) are set to the same as BeautyGAN, i.e.,  $\alpha = 1$, $\beta = 10$, $\gamma = 0.005$ and~$\sigma = 1$. We set~$N_{stand} = 6$ which is the same as the original BeautyGAN, and~$N_{decom} = 9$ for the student network. We first train the student network from scratch for~$80$ epochs till convergence, then the feature loss of CKD is added to the student network and it is trained for another~$60$ epoches. The initial learning rate is~$2 \times 10^{-4}$ and it decays linearly after~$30$ epoches.

\subsection{The Compression Effects}
Table~\ref{tab:Compression Effect} compares the computation and storage of the proposed student network and BeautyGAN. It can be seen that compared with the original BeautyGAN, The proposed student model consumes around~$1/3$ parameters, $1/2$ MACs and~$81.5\%$ inference time.
\begin{table}[htp]
		\caption{Computation and strage comparison between the BeautyGAN and our proposed student model on the Intel(R) Xeon(R) CPU E5-2620 v4 @ 2.10GHz (the proposed CPU).}
		\centering
		\begin{tabular}{lcccc}\bottomrule%\midrule
           Model&\#Params&Model Size&MACs&Inference Time\\\hline%($10^6$)
			BeautyGAN&$9.23M$&$36098KB$&$66.891G$&0.4952s\\
			Proposed&$3.13M$&$12312KB$&$38.269G$&0.4037s\\\bottomrule%\midrule
		\end{tabular}
		\label{tab:Compression Effect}
\end{table}

\begin{figure}[htbp]
\begin{tabular}{cccc}
Non-makeup&Reference&BeautyGAN&Proposed\\
\includegraphics[scale=0.2]{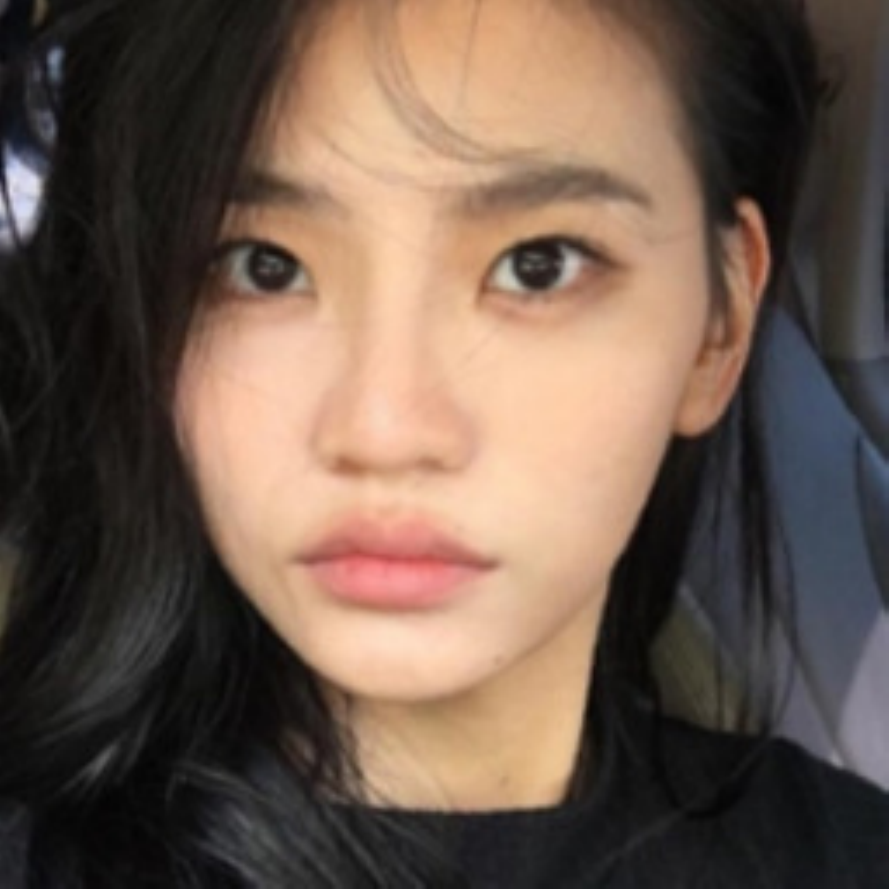}&\includegraphics[scale=0.2]{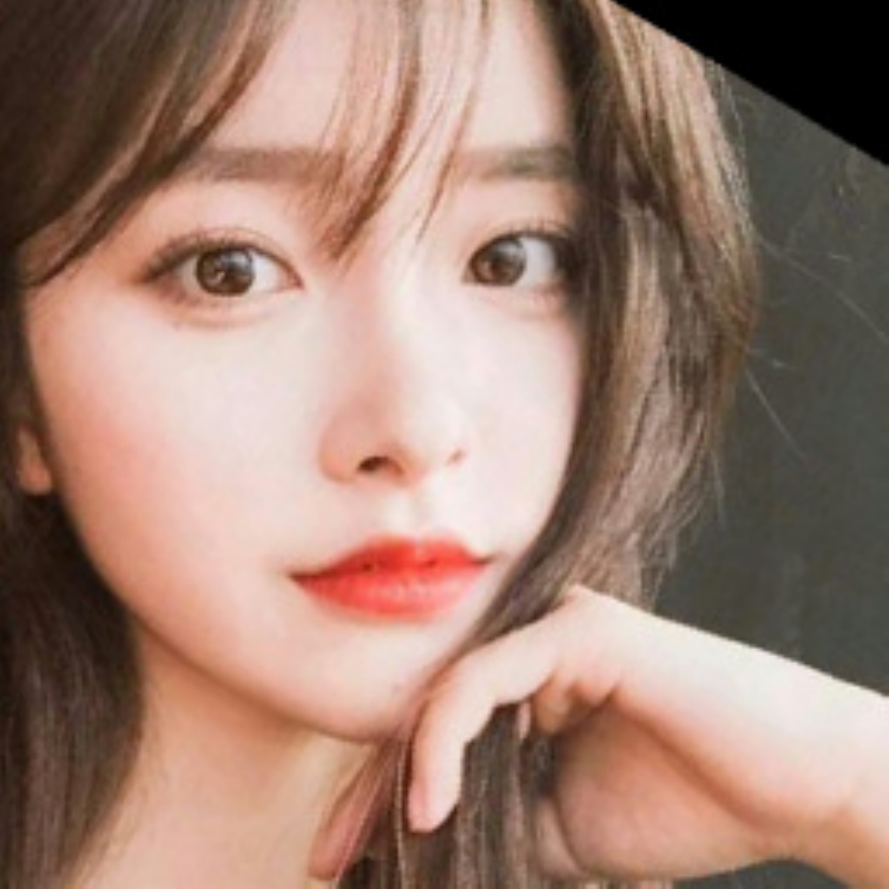}&\includegraphics[scale=0.2]{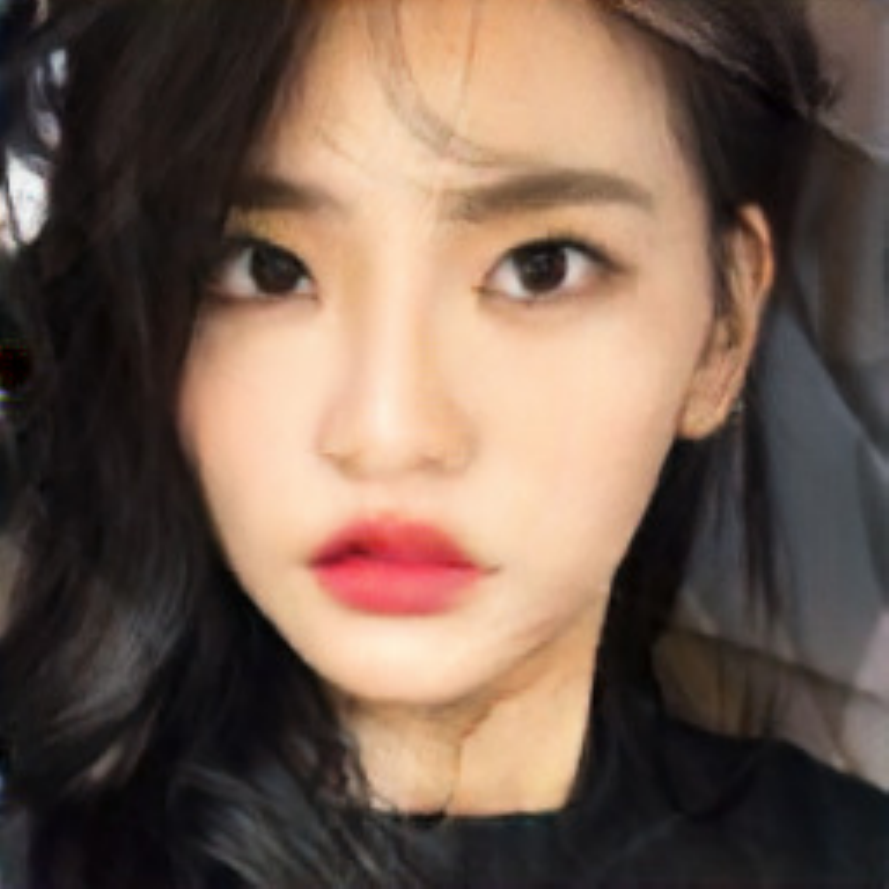}&\includegraphics[scale=0.2]{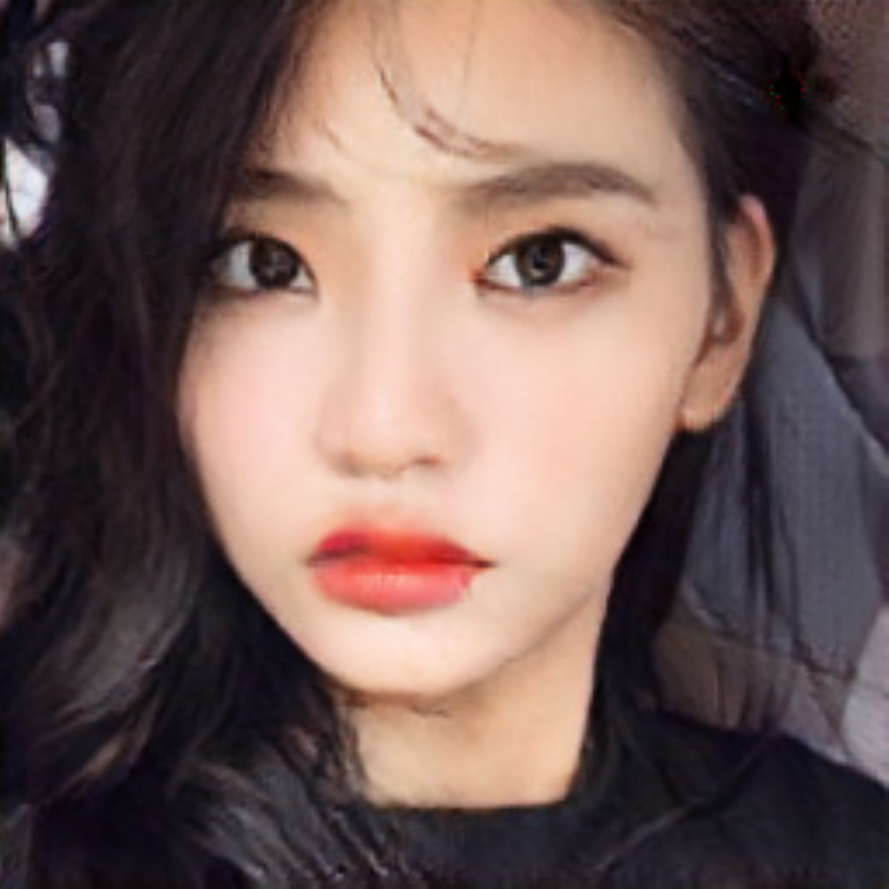}\\
\includegraphics[scale=0.2]{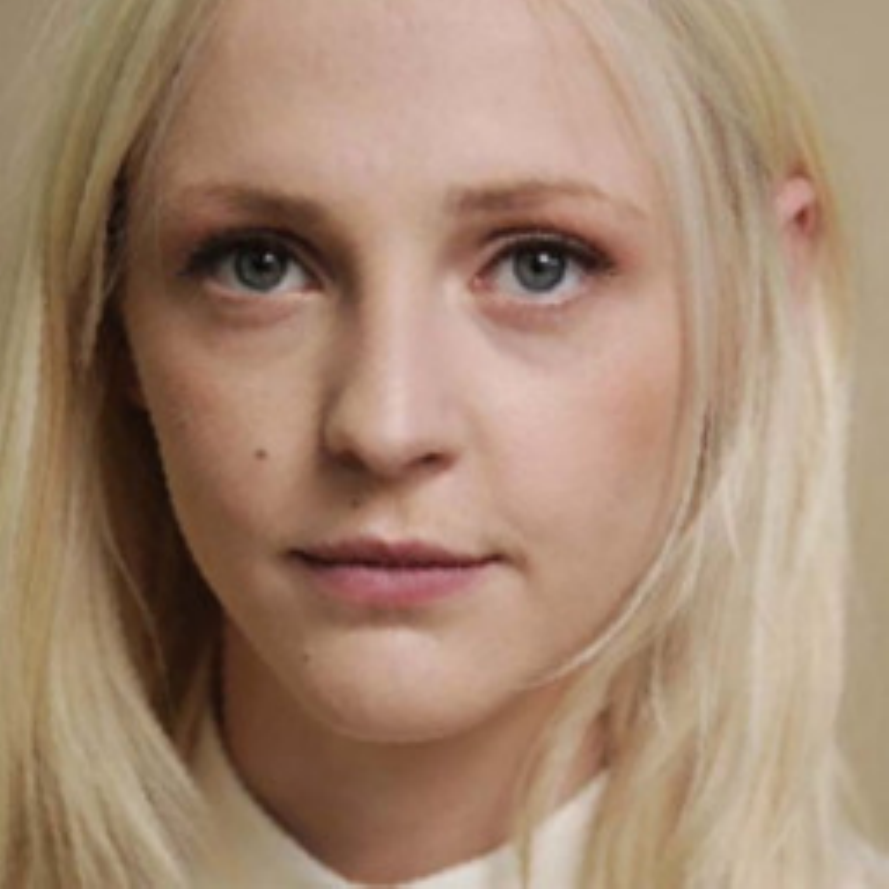}&\includegraphics[scale=0.2]{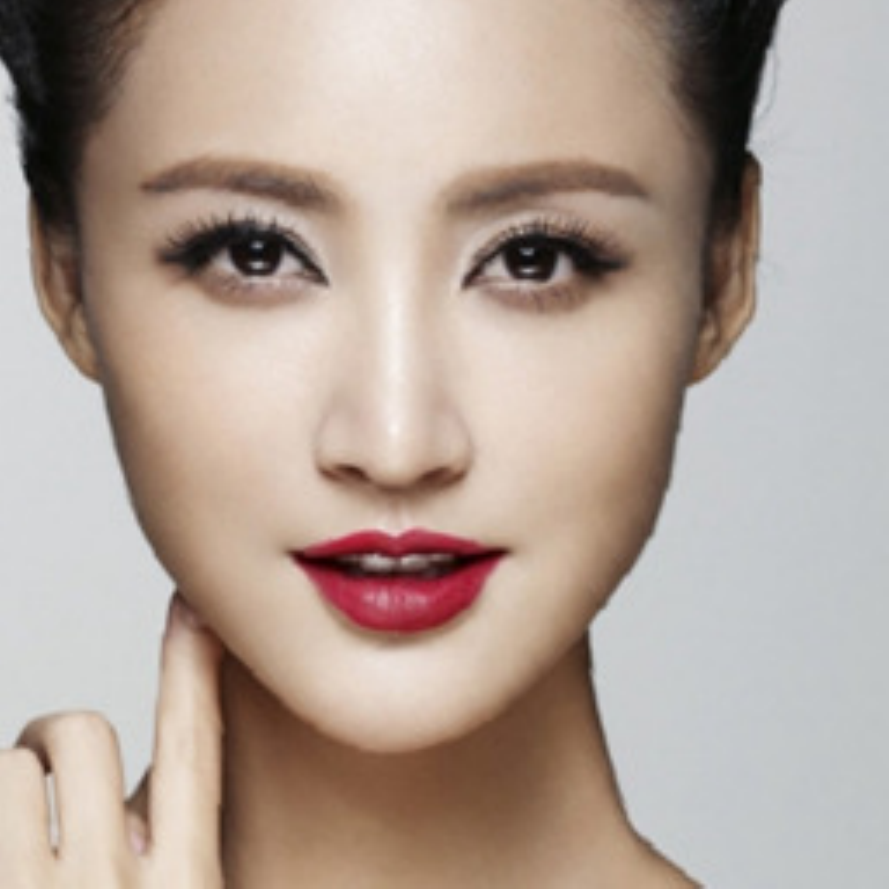}&\includegraphics[scale=0.2]{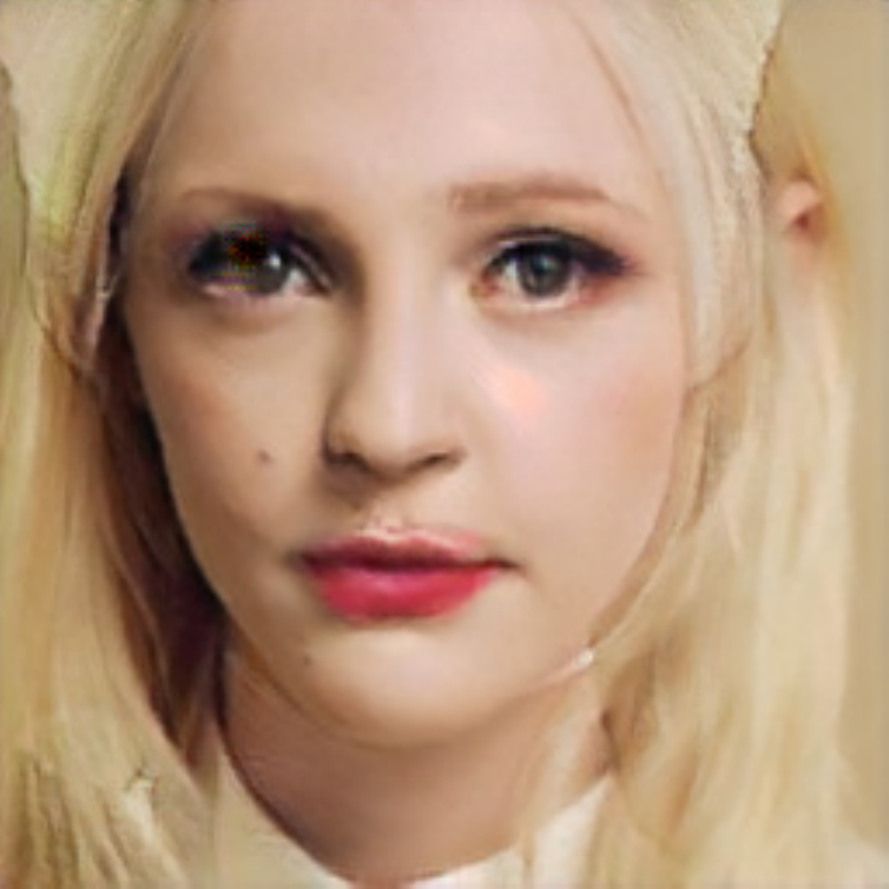}&\includegraphics[scale=0.2]{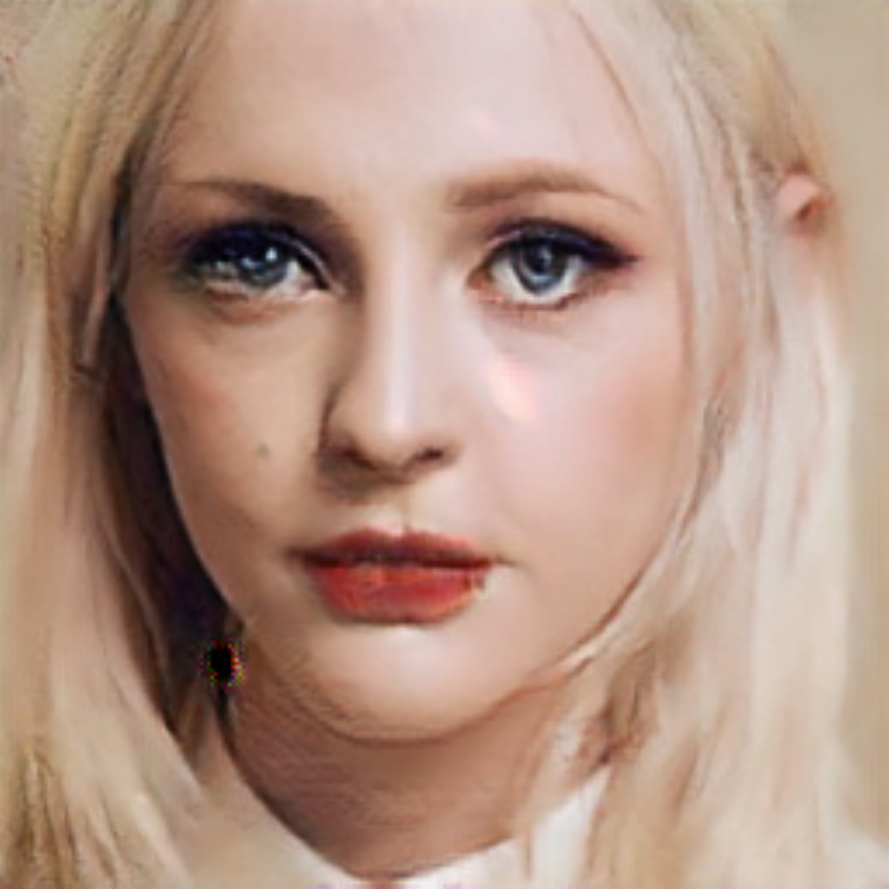}\\
\includegraphics[scale=0.2]{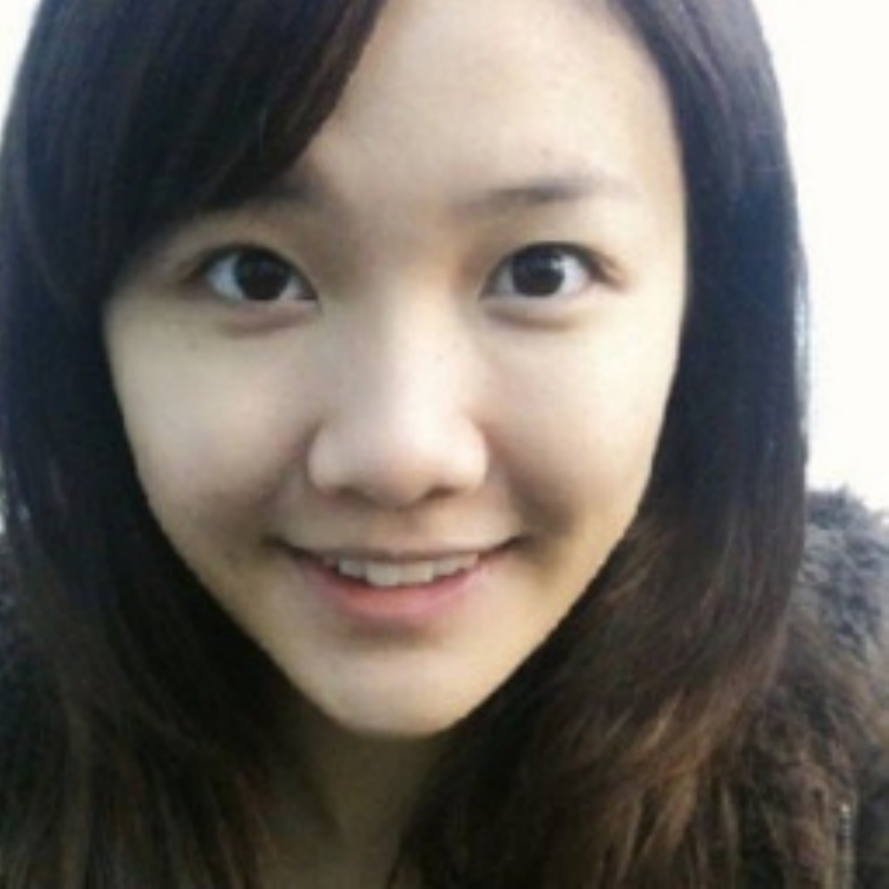}&\includegraphics[scale=0.2]{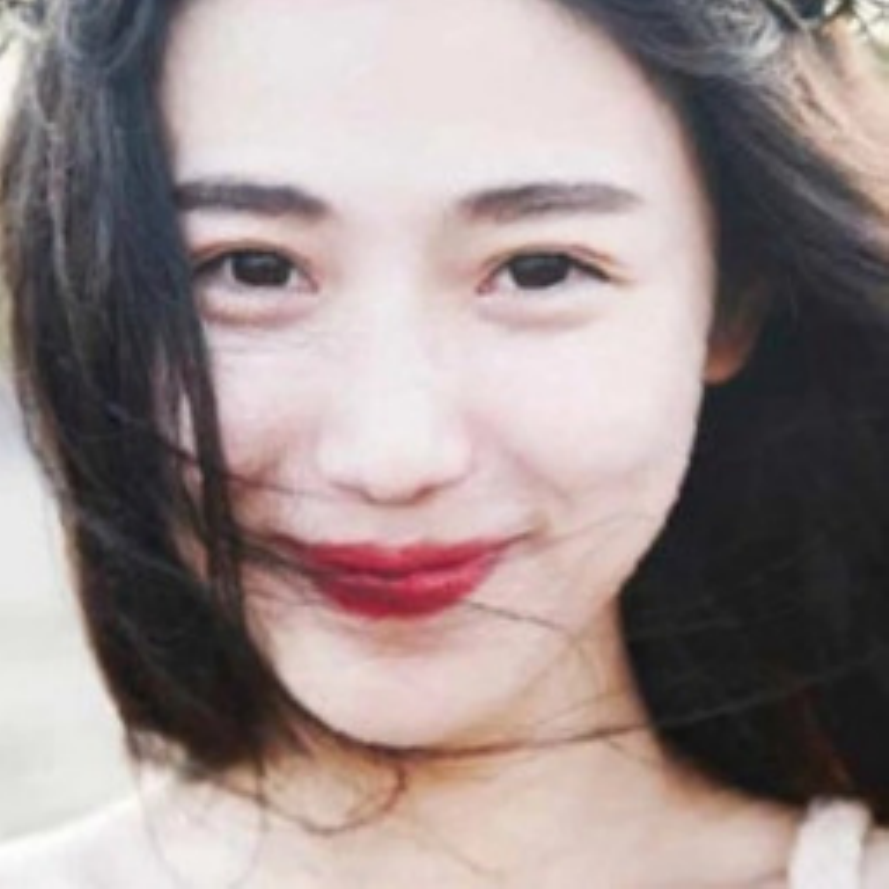}&\includegraphics[scale=0.2]{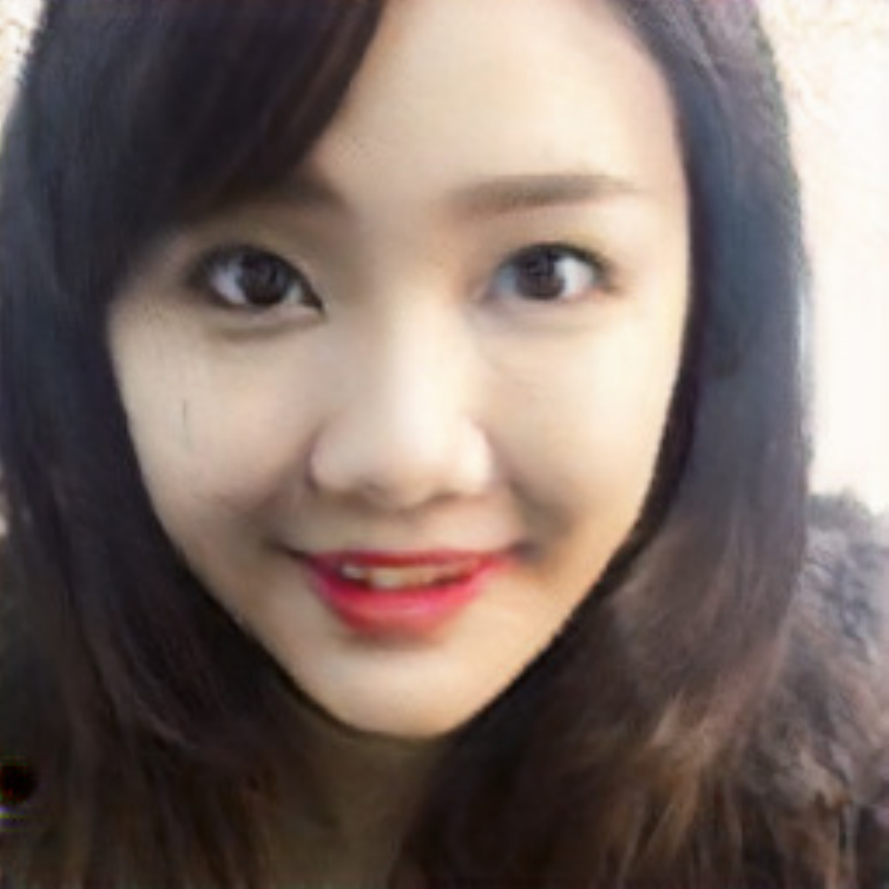}&\includegraphics[scale=0.2]{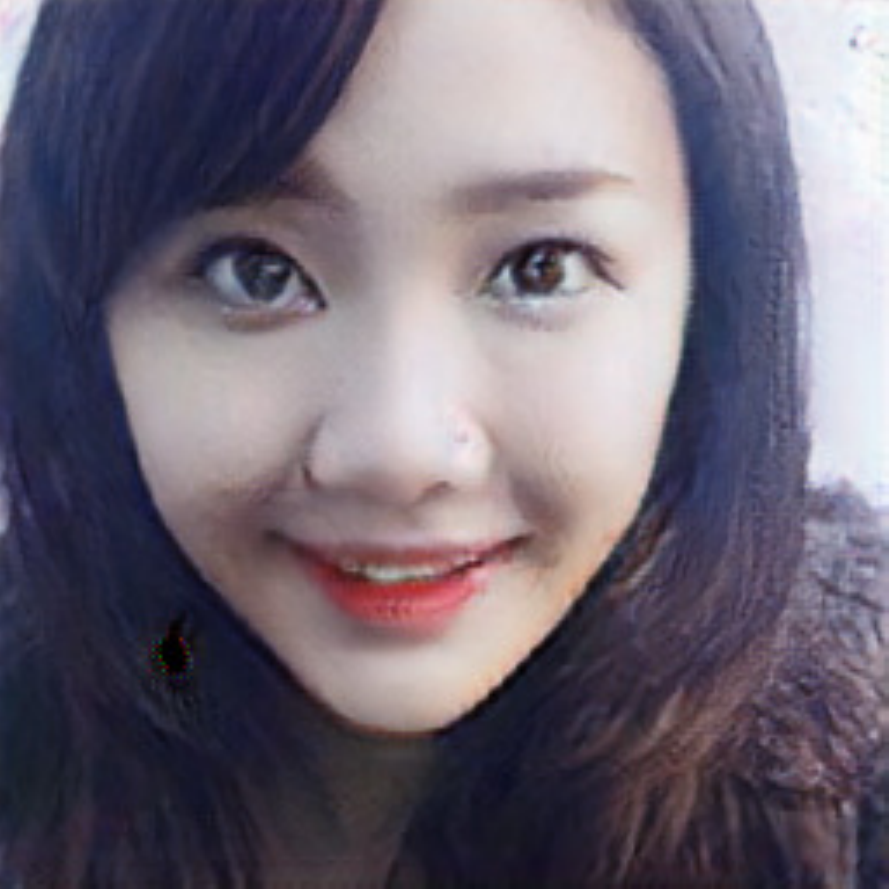}\\
\end{tabular}
\caption{Makeup images generated by BeautyGAN and the proposed student network.}
\label{fig:faces}
\end{figure}

Figure~\ref{fig:faces} shows the makeup images generated by BeautyGAN and the proposed student network, respectively. It can be seen that the perceptual quality of both models is comparable. For some images such as the first and the last rows, makeup images generated by the student network are more similar with the reference in illumination.

To further compare the quality of the generated images, we conduct a user study by randomly selecting~$10$ image pairs, of which one is generated by BeautyGAN, and the other is generated by the proposed student network. Together showing the corresponding non-makeup and reference images, we ask the~$111$ users to choose makeup images which are better in perceptual quality. The result is that of all~$1,110$ votes, there are~$41.08\%$ for BeautyGAN and~$58.92\%$ for the student network, which also indicates that the proposed student network generates comparable even better makeup images than the original BeautyGAN

The perceptual quality can be quantitatively evaluated by two metrics -- the makeup distance~$D_{makeup}$, which is calculated by the \emph{histogram loss} between the makeup and reference images~\cite{LiQiaDonLiuYanZhuLin18}, and the face distance~$D_{face}$, which is the \emph{perceptual loss} calculated by extracting feature maps of makeup faces and non-makeup faces from the $18$th layer of the pretrained VGG16 model~\cite{JohAlaFei16}. For the proposed student network, the makeup distance averaged over all testing images is~$15.09$, compared with~$15.13$ for BeautyGAN. In addition, the average face distance of the proposed student network is~$0.163$, compared with~$0.188$ for BeautyGAN. Thus, the proposed student model performs better for both metrics.

\subsection{Ablation Study}
\begin{figure}[htbp]
\begin{tabular}{cccc}
Non-makeup&Reference&Not Distilled&Distilled\\
\includegraphics[scale=0.2]{1-1inA-eps-converted-to.pdf}&\includegraphics[scale=0.2]{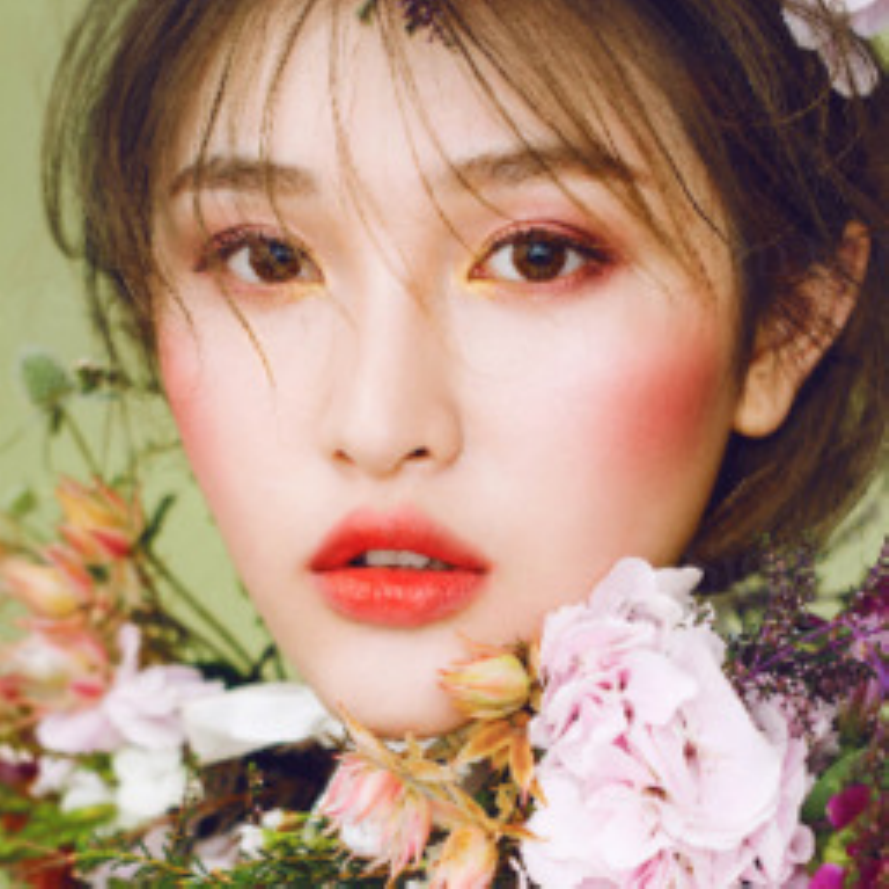}&\includegraphics[scale=0.2]{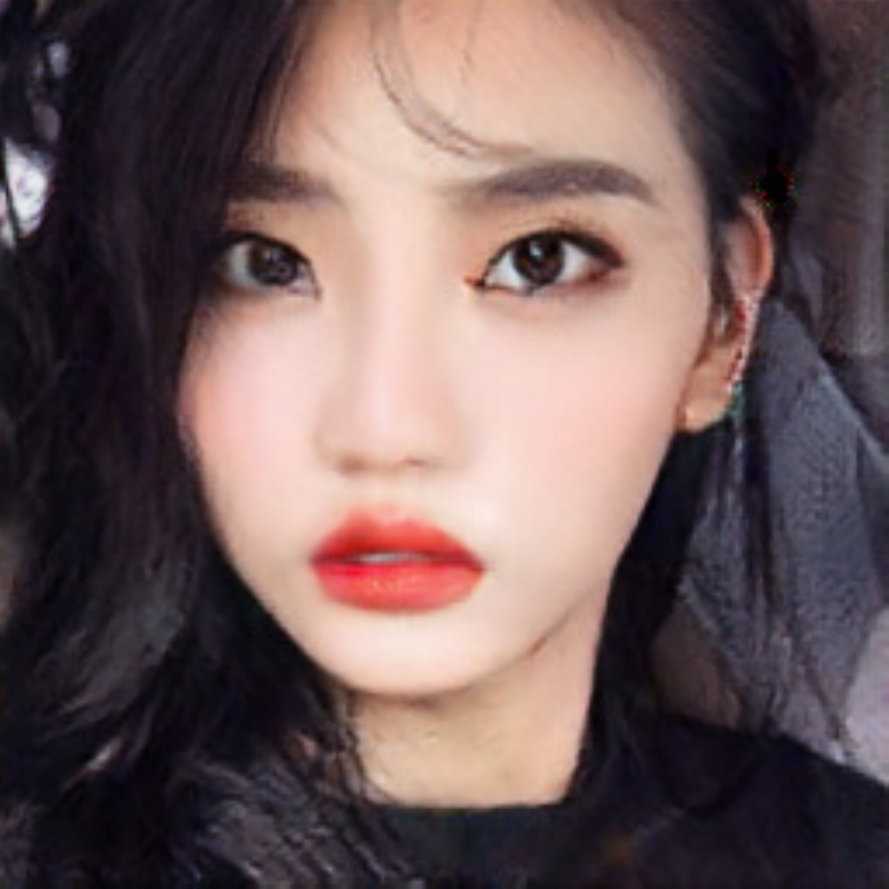}&\includegraphics[scale=0.2]{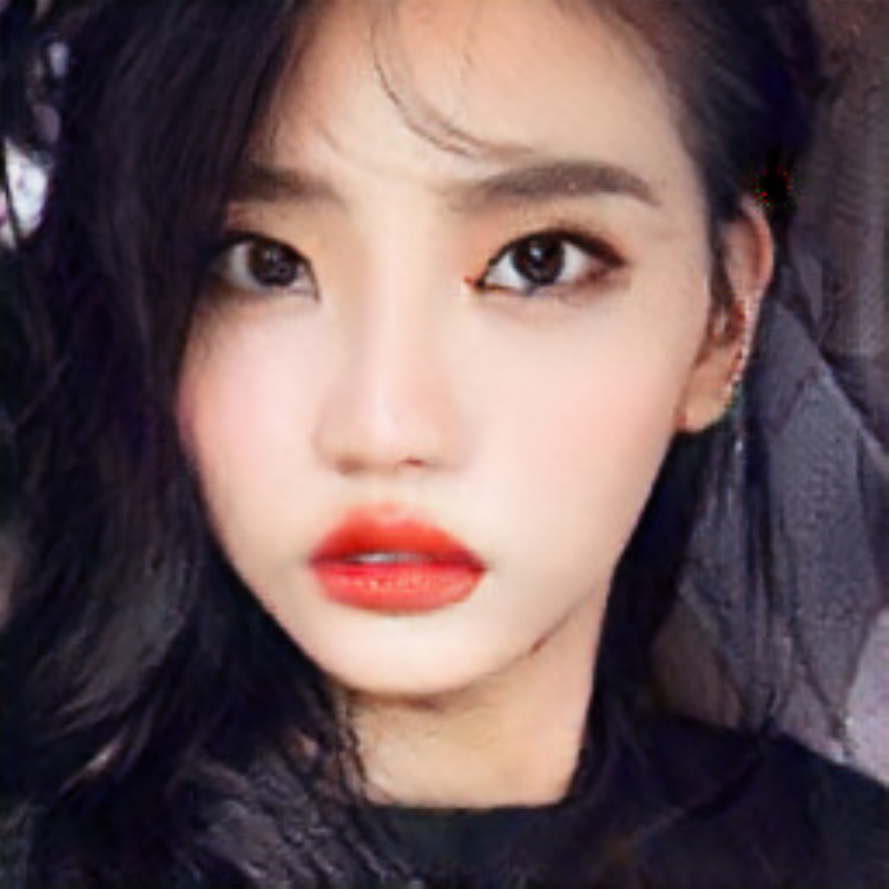}\\
\includegraphics[scale=0.2]{1-1inA-eps-converted-to.pdf}&\includegraphics[scale=0.2]{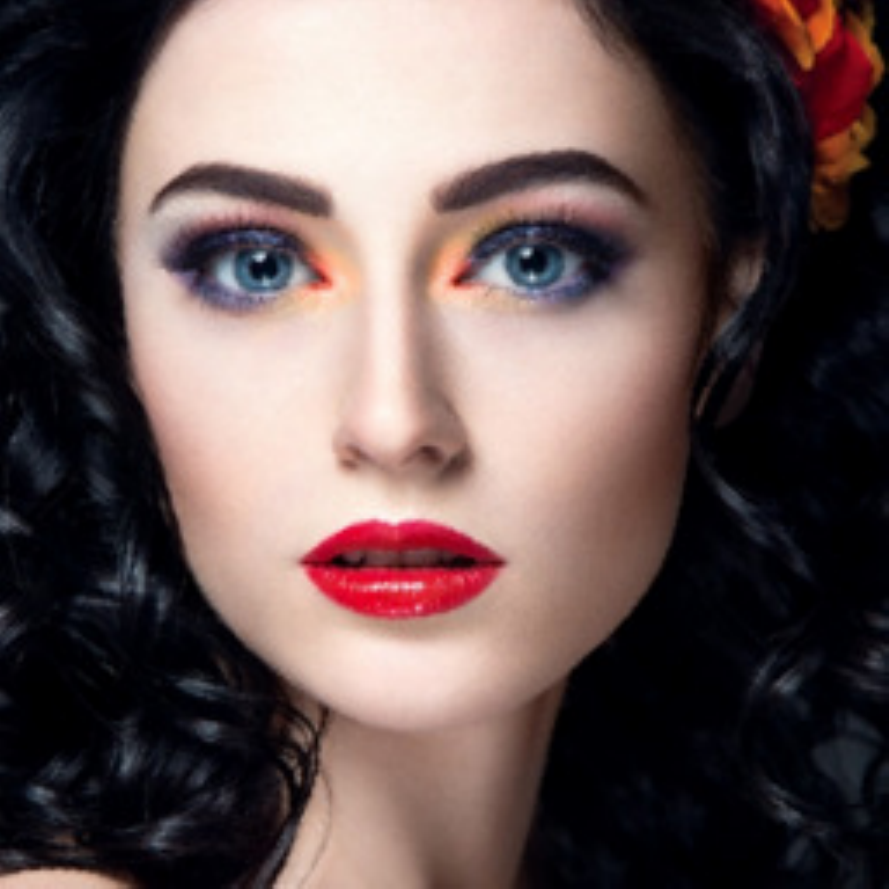}&\includegraphics[scale=0.2]{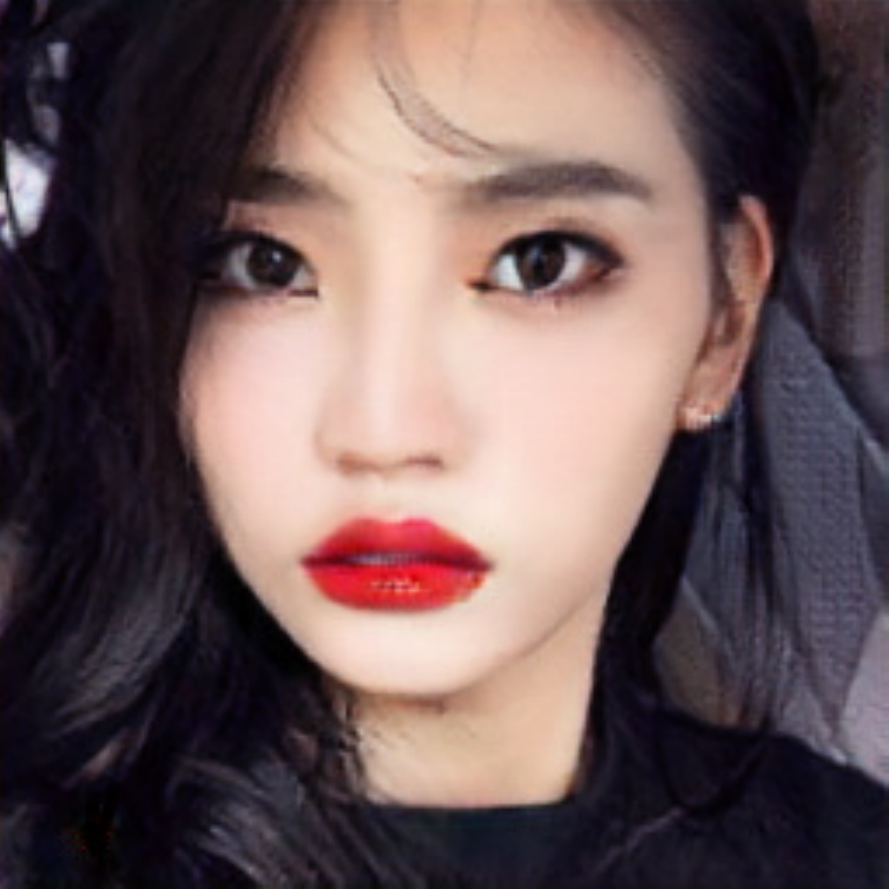}&\includegraphics[scale=0.2]{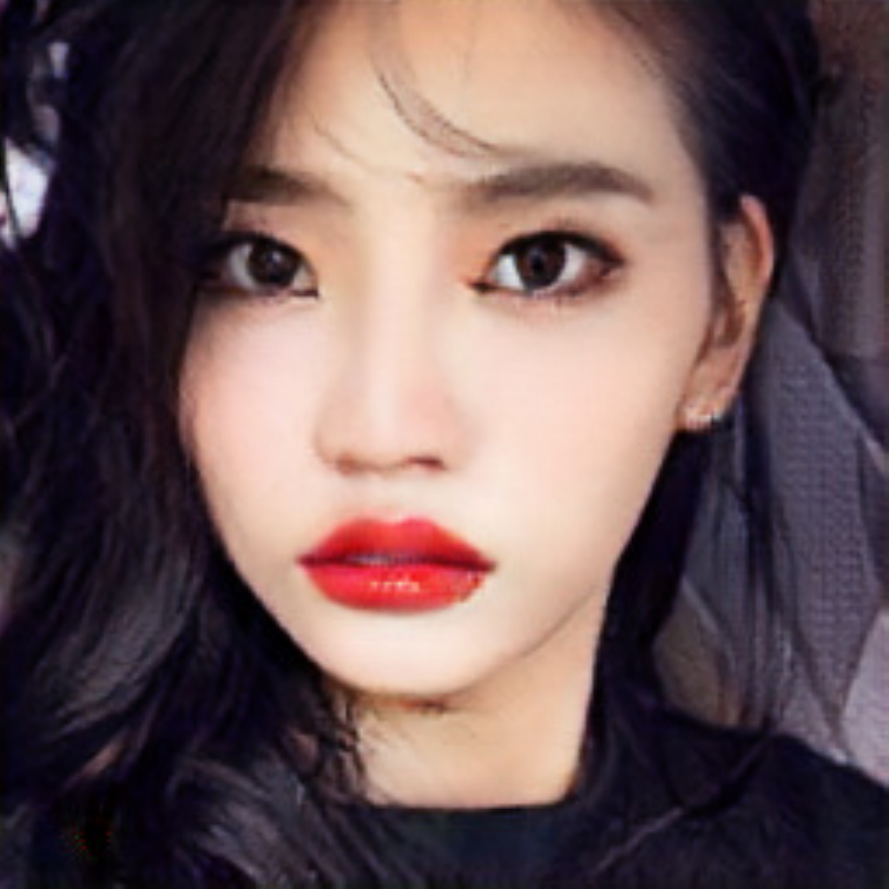}\\
\includegraphics[scale=0.2]{1-1inA-eps-converted-to.pdf}&\includegraphics[scale=0.2]{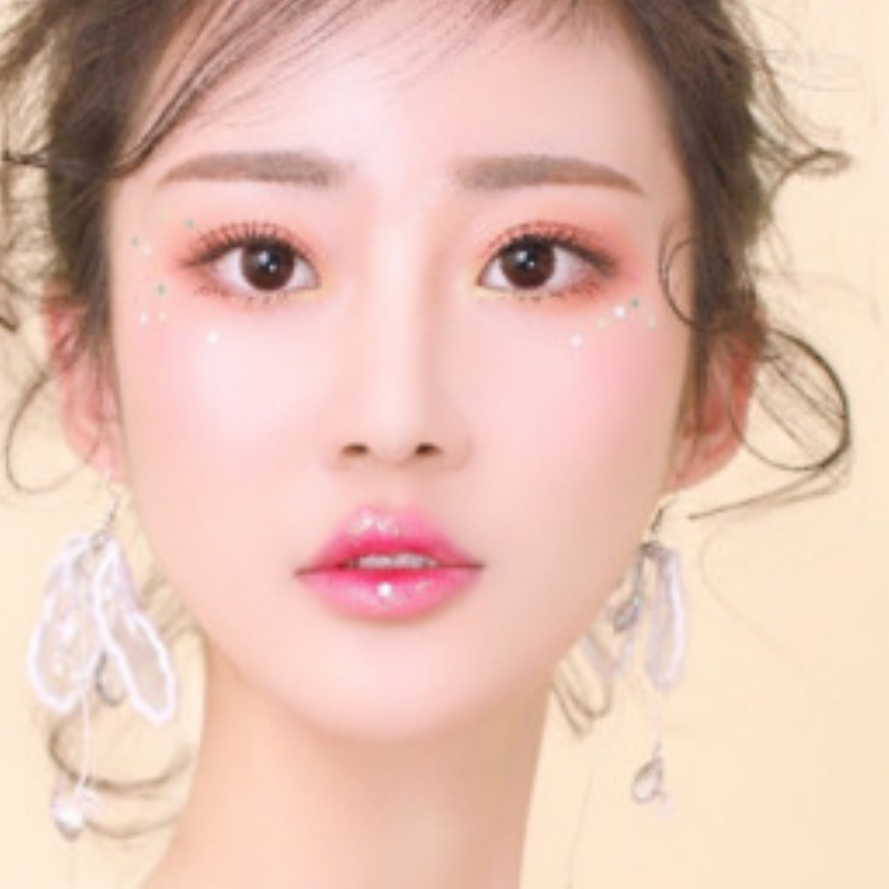}&\includegraphics[scale=0.2]{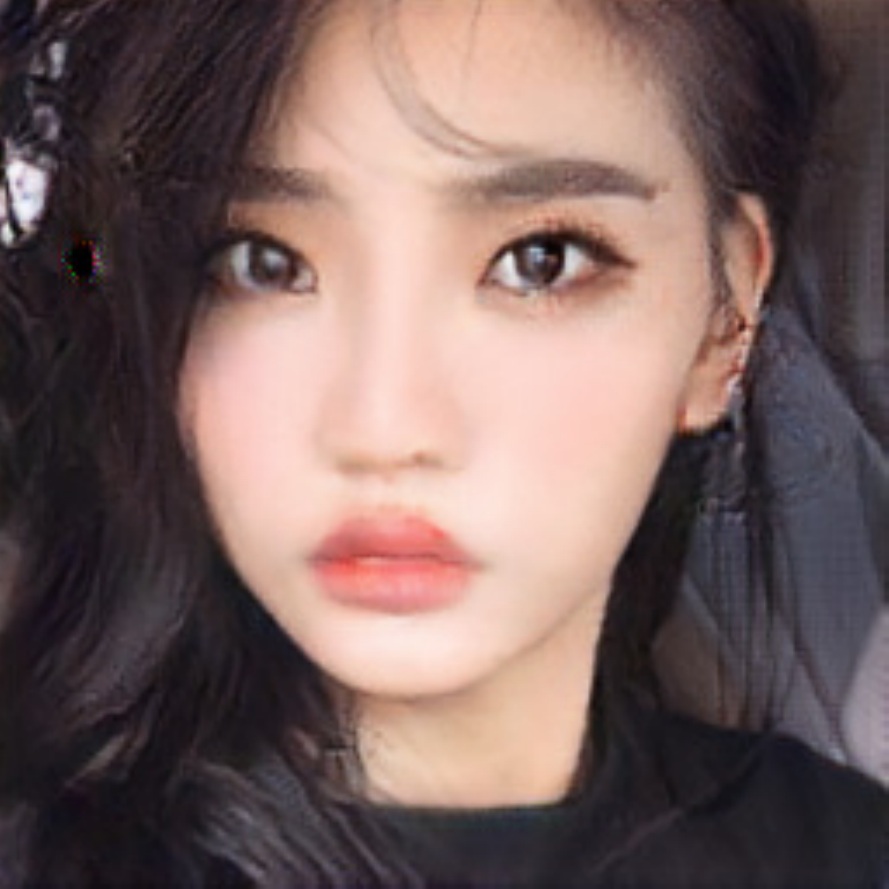}&\includegraphics[scale=0.2]{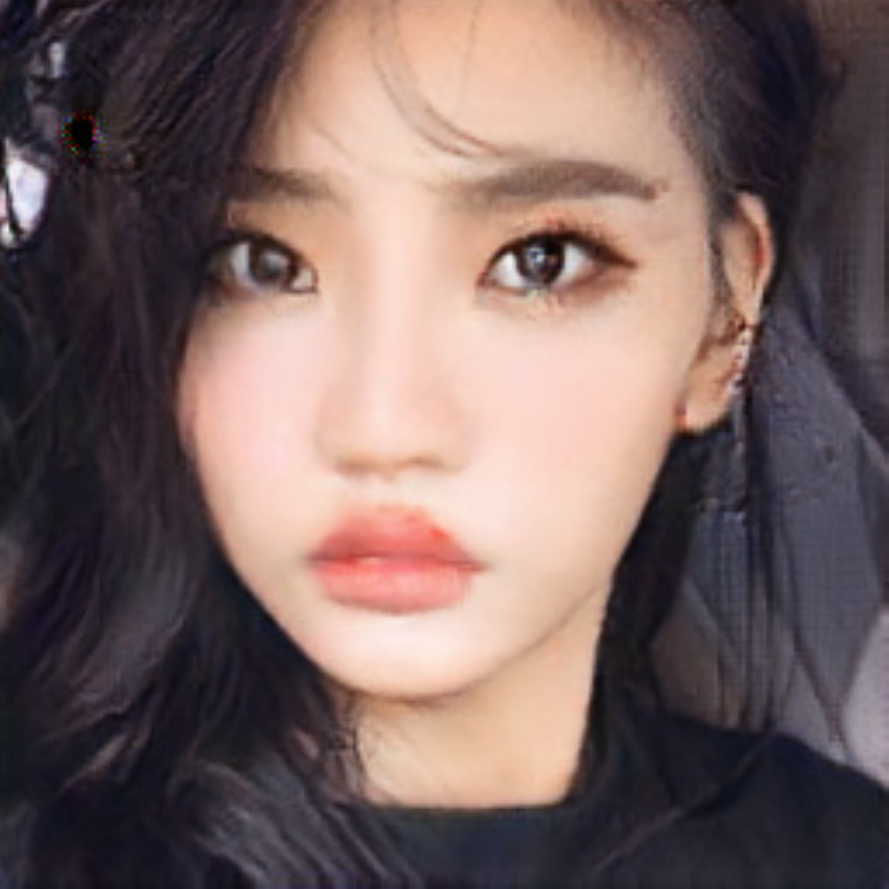}\\
\end{tabular}
\caption{Generated makeup images by student models without/with distillation.}
\label{fig:disillvsdecompose}
\end{figure}

The first ablation study is to compare the performance of the student models without and with the distillation process. Figure~\ref{fig:disillvsdecompose} shows three generated images without/with distillation. It can be seen that both models generate makeup images with similar perceptual quality. Table~\ref{tab:disillvsdecompose} shows~$D_{makeup}$, $D_{face}$ and the inference time averaged over all testing images. It can be seen that the distilled model obtains smaller~$D_{makeup}$ but bigger~$D_{face}$ than the non-distilled model, which also indicates that the performance of these two models is comparable. Because the distilled model uses a smaller encoder to replace the encoder of the original BeautyGAN, it has less inference time than the non-distilled one. This ablation study shows that the main functionality of distillation  is to save computation rather than generate perceptual pleasing images.
\begin{table}[h]
\caption{Comparison of generated images without/with distillation by two distance metrics and inference time on the proposed CPU averaged over~$234$ testing images.}
\centering
\begin{tabular}{lcc}\bottomrule%\midrule
       &Not Distilled &Distilled\\\hline%($10^6$)
$D_{makeup}$&$14.3322$&$\textbf{14.3171}$\\
$D_{face}$&$\textbf{0.1218}$&$0.1245$\\
Inference Time(s)&$0.4424$&$\textbf{0.3657}$\\\bottomrule%\midrule
\end{tabular}
\label{tab:disillvsdecompose}
\end{table}

Our second ablation study is to obtain the optimal number of residual modules. Table~\ref{tab:res9_reason} is the  comparison when student models have different numbers of decomposed residual blocks. It can be seen that when~$N_{decom}=9$, it obtains the smallest~$D_{face}$ value and the inference time is also less than $N_{decom}=10$. %The user study also indicate that $N_{decom}=9$ is perceptually better than~$N_{decom}=6,8,10$, but space precludes further details.
Thus, we choose~$N_{decom}=9$ for our experiments.
\begin{table}[htbp]
		\caption{Quantitative study on the proposed CPU when student models have different numbers of residual blocks.}
		\centering
		\begin{tabular}{lcccc}\bottomrule%\midrule
            $N_{decom}=$     &$6$&$8$&$9$&$10$\\\hline%($10^6$)
			$D_{makeup}$&$16.5514$&$16.5628$&$16.5538$&$\textbf{16.4596}$\\
			$D_{face}$&$0.1148$&$0.1401$&$\textbf{0.1072}$&$1.7742$\\
           Inference Time(s)&$\textbf{0.3926}$&$0.4233$&$0.4424$&$0.4535$\\\bottomrule%\midrule
		\end{tabular}
		\label{tab:res9_reason}
\end{table}

\section{Conclusion}
To save computation and storage of facial makeup transfer networks, we introduce a two-step compression method including collaborative knowledge distillation and kernel decomposition for BeautyGAN. Extensive experiments show the effectiveness of the proposed facial makeup transfer network.

\section{ACKNOWLEDGMENT}
This work is supported by the Natural Key R\&D Program of China (Grant No. 2017YFB1002400) and Natural Science Foundation of Zhejiang Province (Grant No. LY16F010004).

\bibliographystyle{IEEEtran}
\bibliography{references}

\end{document}